\newif\ifcomments
\documentclass{llncs}
\usepackage{graphicx}
\usepackage{amsmath,amssymb} % define this before the line numbering.
\usepackage{color}
\usepackage[width=122mm,left=12mm,paperwidth=146mm,height=193mm,top=12mm,paperheight=217mm]{geometry}
\usepackage{times}
\usepackage{epsfig}
\usepackage{graphicx}
\usepackage{amsmath}
\usepackage{amssymb}
\usepackage{tikz}
\usepackage{float}
\usepackage{subcaption,multirow}
\usepackage[pagebackref=true,breaklinks=true,letterpaper=true,colorlinks,bookmarks=false]{hyperref}
\usepackage{xspace}

\usepackage{xcolor}
\newcommand*{\affmark}[1][*]{\textsuperscript{#1}}
\ifcomments
\newcommand\todos[1]{\textcolor{red}{#1}}
\newcommand\newtodos[1]{\textcolor{blue}{#1}}
\newcommand{\shaodi}[1]{\textcolor{green}{Shaodi: [#1]}}
\newcommand{\etc}[1]{~\textit{etc.}}
\newcommand{\etal}[1]{~\textit{et~al.}}
\usepackage[normalem]{ulem}
\else
\newcommand\todos[1]{}%{\textcolor{red}{#1}}
\newcommand\newtodos[1]{}%{\textcolor{blue}{#1}}
\newcommand{\shaodi}[1]{}%{\textcolor{green}{Shaodi: [#1]}}
\newcommand{\etc}[1]{}%{~\textit{etc.}}
\newcommand{\etal}[1]{}%{~\textit{et~al.}}
\newcommand{\sout}[1]{}
\fi
\usetikzlibrary{positioning,spy}

\makeatletter
\DeclareRobustCommand\onedot{\futurelet\@let@token\@onedot}
\def\@onedot{\ifx\@let@token.\else.\null\fi\xspace}

\def\eg{\emph{e.g}\onedot} 
\def\ie{\emph{i.e}\onedot} 
 
\def\etc{\emph{etc}\onedot} 
 
\def\etal{\emph{et al}\onedot}
\makeatother

\begin{document}
%\renewcommand\thelinenumber{\color[rgb]{0.2,0.5,0.8}\normalfont\sffamily\scriptsize\arabic{linenumber}\color[rgb]{0,0,0}}
%\renewcommand\makeLineNumber {\hss\thelinenumber\ \hspace{6mm} \rlap{\hskip\textwidth\ \hspace{6.5mm}\thelinenumber}}
% \linenumbers
%\pagestyle{headings}
%\mainmatter
%\def\ECCV18SubNumber{}  % Insert your submission number here

\title{Semantic Single-Image Dehazing} % Replace with your title

%\titlerunning{ECCV-18 submission ID \ECCV18SubNumber}

%\authorrunning{ECCV-18 submission ID \ECCV18SubNumber}

\author{
  Ziang Cheng\affmark[1,2] \and
  Shaodi You\affmark[1,2] \and
  Viorela Ila\affmark[1] \and
  Hongdong Li\affmark[1]
}
\institute{Australian National University \and Data61 CSIRO, Australia}

\maketitle

\begin{abstract}
Single-image haze-removal is challenging due to limited information contained in one single image. Previous solutions largely rely on handcrafted priors to compensate for this deficiency. Recent convolutional neural network (CNN) models have been used to learn haze-related priors but they ultimately work as advanced image filters. In this paper we propose a novel semantic approach towards single image haze removal. Unlike existing methods, we infer color priors based on extracted semantic features. We argue that semantic context can be exploited to give informative cues for (a) learning color prior on clean image and (b) estimating ambient illumination. This design allowed our model to recover clean images from challenging cases with strong ambiguity, \eg saturated illumination color and sky regions in image. \sout{\todos{how powerful the method is}} In experiments, we validate our approach upon synthetic and real hazy images, where our method showed superior performance over state-of-the-art approaches, suggesting semantic information facilitates the haze removal task. 
%\keywords{single image, dehaze, semantic, haze/fog removal, deep learning}
\end{abstract}

\section{Introduction}

%why dehaze
Images taken in hazy/foggy weather are generally subject to visibility degradation caused by particle-scattered light, this includes color shifting, contrast loss and saturation attenuation, \etc, which may in turn jeopardize the performance of high-level computer vision tasks, \eg object recognition/classification, aerial photography, autonomous driving and remote sensing. 

%brief literary review
Existing dehazing algorithms follow a well-received particle model~\cite{narasimhan2003contrast}, which correlates scene structure and haze-free image under given hazy inputs. Early research make use of multiple images of the same scene taken from different angles/positions to recover structural information and consequently, haze concentration. Single image dehazing, on the other hand, is an extremely ill-posed problem. The challenge arises from the fact that a single hazy image is not informative on scene structure nor clean scene, whereas one is needed to infer the other, causing an ambiguity in clean image estimation. Generally, traditional methods \cite{he2011single,zhu2015fast,berman2016non}\sout{\shaodi{cite a few here}} explicitly leverage priors (or constraints) on both fronts: local depth coherence is often assumed, and so is one or more color priors. Recently, several CNN models were proposed for dehazing \cite{cai2016dehazenet,ren2016single,li2017aod}\sout{\shaodi{cite a few here}}, and have been found on comparable level with prior arts in term of performance. These models are generally light-weighted, and essentially leverage mostly lower level features as dehaze is still considered as an image processing problem.

%problems with prior arts
In a word, existing methods are solely relying on handcrafted physical or low-level priors. While these empirical priors work generally well, one can easily make counterexamples which violate their assumptions (\eg bright surface, colored haze). A well-observed problem with this is that the restored image tends to be over-saturated and has incorrect tone. This problem is in general unsolvable based on the insufficient low-level information alone. However, human users can easily tell the naturalness of color, \eg whether the tree is too green or the sky is too blue. Such feeling is not from any low-level priors but rather because of the semantic prior that we humans have a knowledge about.\\

\begin{figure}[!t]
\hspace{-1.5cm}
\includegraphics[page=1, width=1.25\textwidth, trim={0cm 8.2cm 0cm 8.2cm}, clip]{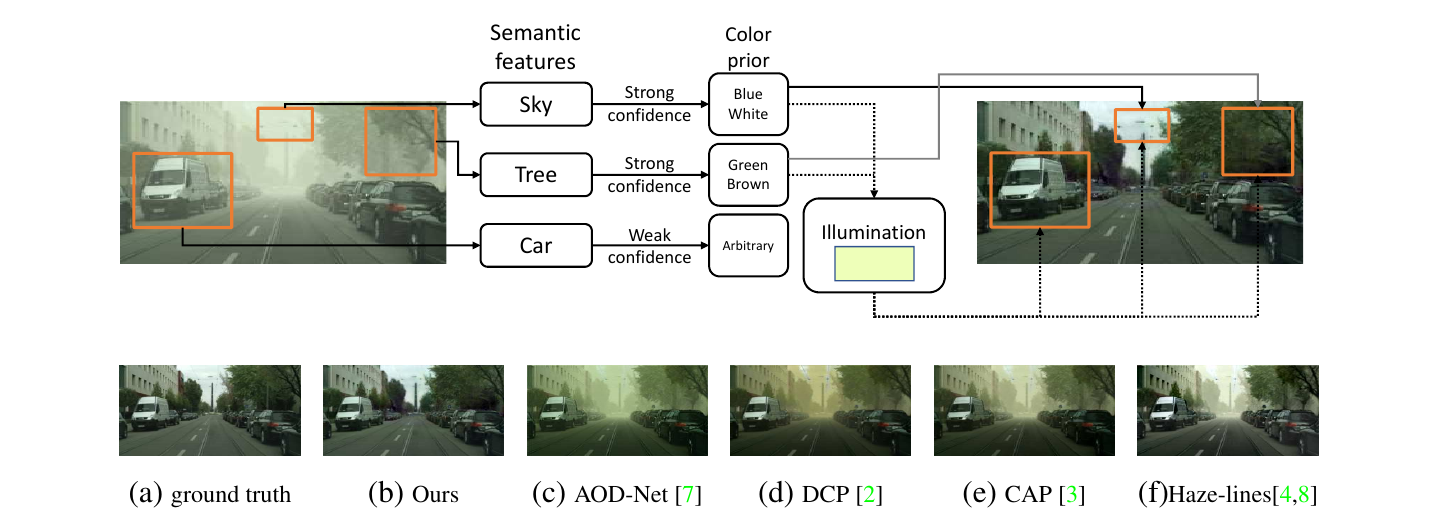}

\caption{The proposed semantic solution towards dehazing: we rely on semantic priors to provide additional information that is otherwise hard to obtain. This extra knowledge can be helpful with predicting objects' true color as well as ambient illumination.}\label{fig:intro}
\sout{\shaodi{The caption can be more detailed, the confidence of color prior is not mentioned.}
\shaodi{Looks like some JPG artifacts in the sky region, can we clean it? \\It's overhead lines}}
\end{figure}

%intro to our methods
In this paper we introduce semantics-based dehazing, a novel method that uses semantic information to provide additional guidance for inferring clean image. Semantic clues have seen success in other `low-level' applications, \eg color constancy~\cite{hu2017fc} and image filtering~\cite{yang2016semantic}. Here we propose a fully end-to-end convolutional neural network (CNN) that learns the correlation between semantics and objects' natural color from training samples, and infer the clean scene and illumination color based on learned semantic features. As such, for object class of medium or strong semantic color prior (\eg sky is blue and vegetations are green), the semantics provides informative cues on the object's true color, and the clean scene and ambient illumination can be learned with high confidence; for objects of medium or weak semantic prior, the true color can be predicted with, for example, low-level priors and ambient illumination estimation from other strongly confident objects. The conceptual idea behind our approach is illustrated in Figure~\ref{fig:intro}.

The main contribution of this paper is that we are the first to explicitly exploit high-level features to provide informative color priors for single image haze removal problem. We found that our approach is robust against extreme settings (bright surfaces, severe color shifting, saturated atmospheric light, sky regions \etc) which impose major challenges on previous methods. \sout{\todos{make sure that in the experiments section you exemplify all/most of those}} We show in experiments that the proposed model obtains state-of-the-art results on RGB-D testsets with synthetic haze. As the model is trained on street scenes, we also test our model on real world hazy scenes of similar semantic classes, where the model shows comparable results with state-of-art methods. \sout{\todos{this is probably another place to mention that the generality of the method is constrained by the semantic classes}}

\section{Related work}

\subsection{Atmospheric scattering model}

%Visibility degradation caused by haze and fog is the result from particles and/or water droplets scattering light rays in all directions. With the presence of haze or fog, a portion of true object radiance/reflection is scattered by particles, leading to transmission attenuation of true scene. On the other hand, the ambient illumination causes brightness and color shifting. 

Following~\cite{narasimhan2003contrast}, hazy imagery can be seen as the linear combination of true object color and ambient illumination (Figure~\ref{fig1}), hence the equation
\begin{equation}
  \label{phyxmdl}
  I(x) = J(x)\, t(x) + A\,\big(1-t(x)\big)\text{,}
\end{equation}
where $I(x)$ is the hazy image value for pixel $x$, $J(x)$ is the corresponding clean image\sout{\todos{ is this the radiance measured in $\frac{W}{m^2\: sr}$ ? No, radiance means clean scene, \ie rgb color}}, $A$ is the color of ambient illumination, and $t(x)\in(0,1]$ represents the transmission. Assuming the haze is uniformly distributed in space, the transmission $t(x)$ is defined as
\begin{equation}
  \label{trnsdef}
  t(x)=e^{-\beta\,d(x)}\text{,}
\end{equation}
where $d(x)$ is the object distance from camera center \sout{\todos{is this the depth map (z-map) or the distance to the objects}} and $\beta$ a non-negative scattering coefficient related to haze particles. 

\begin{figure}[!t]
\centering
\includegraphics[width=.8\textwidth, trim={0 0 0 5cm}, clip]{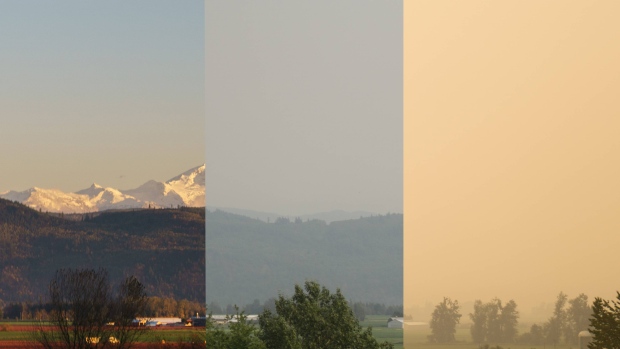}
\caption{Mount Baker from distance in different weather (taken from Internet). Clean image (left) and two different illumination colors (middle and right) are stitched together.}\label{fig:airlight}
\end{figure}

Sometimes it is assumed that $A$ is a bright gray/white color~\cite{zhu2015fast,ren2016single,cai2016dehazenet}. However, with some particles or specific lighting conditions, $A$ may take other color as well (\eg yellow/red), as illustrated in Figure~\ref{fig:airlight}.

\begin{figure}[htbp]
\centering
\vspace{-.5cm}
\includegraphics[width=0.8\textwidth, trim={0cm 6cm 3cm 5cm}, clip]{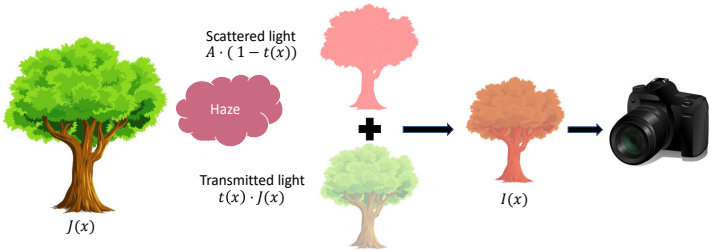}
\caption{Atmospheric scattering model. The captured image is a convex combination of ambient illumination and clean scene.}\label{fig1}
\shaodi{we need a beautiful image}
\end{figure}

%Hazy imagery can be seen as the linear combination of true object radiance and ambient illumination (Figure~\ref{fig1}). Under clear weather (\ie $\beta=0$), the transmission is $1$, thus the captured image represents true scene radiance. On the other hand, when depth is large enough (\eg in sky region), transmission approaches $0$, and the corresponding pixel value is reduced to ambient illumination color.

\subsection{State-of-the-art for haze removal}\label{sota}
\sout{\shaodi{This subsection is too long}}
%difficulty in dehaze and general solution
%Single image haze removal is intrinsically ill-posed as only the hazy image $I(x) $ is given, while both transmission $t(x)$ and radiance $J(x)$ are unknown, \ie there are linearly more unknowns than equations in \eqref{phyxmdl}. Based on empirical or statistical observation, existing non-learning methods generally rely on handcrafted prior assumptions to impose constraints on search space. Typically, transmission is encouraged to be static or smooth in a small image patch based on depth coherence assumption. And color priors are applied on hazy image or the underlying true radiance. 

%state-of-arts
While there have been many developments in single image dehazing (\eg maximal local contrast~\cite{tan2008visibility}, atmospheric light recovery~\cite{sulami2014automatic}, learning framework for haze features~\cite{tang2014investigating}, color-line~\cite{Fattal2014}, artifacts removal for compressed hazy image and video~\cite{chen2016robust}), in this section we will only list some of the most prominent or recent ones. For a comparative survey of other existing dehazing algorithms, we refer the reader to \cite{li2017haze}.

\emph{Dark channel prior} (DCP)~\cite{he2011single} is based on the observation that at least one of RGB channels of real world objects often has a very small value. Under this assumption, the color shifting caused by ambient illumination can be obtained by its dark channel, hence the transmission. %Meng \etal~\cite{meng2013efficient} further exploit dark channel for a lower bound on transmission. Dark channel prior works generally well with natural images, albeit it does not account for the occurrence of bright surfaces, \eg snow and ice, light building walls, \etc, which may be mistakenly treated as haze dense region and the results are often over-saturated. 
\emph{Color attenuation prior} (CAP)~\cite{zhu2015fast} by Zhu \etal creates a linear model correlating scene depth and the difference between local saturation and brightness. %They observed that concentration of haze causes both brightness increase and saturation loss.%, and they learned the model parameters with supervision. 
%However, the reason for linearity is not fully explained.
\emph{Haze-line}~\cite{berman2016non,7951489} assumes that real world images have distinct colors so that image pixels form clusters in RGB space. With the presence of haze, the clusters are shifted towards ambient illumination based on transmission forming so-called haze-lines.
%deep learning methods
Recently some deep learned methods have also been proposed.
\emph{DehazeNet}~\cite{cai2016dehazenet} employs an end-to-end fully convolutional network (FCN)~\cite{long2015fully} to learn the scene transmission. However, the network is trained on small image patches with constant transmission per patch, and does not factor in non-local features. 
\emph{MSCNN Dehaze}~\cite{ren2016single} propose a multi-scale CNN for learning a coarse transmission, and relies on another CNN in the pipeline for refining it. 
\emph{AOD-Net}~\cite{li2017aod} is the first end-to-end model to directly produce clean images, and has been found to boost the performance for high-level vision tasks under hazy weather conditions. 
Yang \etal~\cite{yang2018towards} design three separate networks to generate the clean image, ambient illumination and transmission respectively and use an adversarial network for semi-supervised learning.\sout{\todos{do we need to compare with this one}}

\sout{\todos{missing a concluding sentence to frame the current contribution}}
As image dehazing is generally considered a low-level task, existing dehaze algorithms seek priors from either empirical observations or the physical haze model. While deep learned technologies do not explicitly assume such knowledge, they are light-weighted and are ultimately designed to learn low-level haze-related features.

\subsection{Datasets for single image dehazing}

%review of current dehaze datasets
Since it is difficult to take images under different weather conditions while keeping other scene settings unchanged, currently there is no dataset offering a large number of real world hazy images and the corresponding clean images. 

Previous hazy datasets uniformly synthesize haze on RGB-D images based on \eqref{phyxmdl}. Due to the difficulty of collecting depth map in outdoor settings, most RGB-D \newtodos{are those obtained using RGBD sensors, Kinect? Yes} datasets contain indoor scenes only (\eg D-Hazy dataset~\cite{ancuti2016dhazy}, which use images and depth maps from NYU~\cite{silberman11indoor} and Middlebury~\cite{zbontar2016stereo,scharstein2007learning,scharstein2003high} datasets). Depth maps for outdoor scenes are in general less accurate and are obtained by (a) view disparity from stereo cameras (\eg FoggyCityscapes~\cite{foggy}) or (b) monocular image depth estimation (\eg RESIDE~\cite{li2017reside} used \cite{liu2016learning} to generate depth maps). Apart from that, all existing synthetic hazy datasets use illumination color close to grayscale, the only exception being Fattal's dataset~\cite{Fattal2014}, which selects sky color as illumination color. %Note that most existing CNN dehaze models are trained and tested with indoor dataset (\eg NYU/NYU2 and Middlebury), the only exception being DehazeNet~\cite{cai2016dehazenet} which is trained on small image patches from Internet with of constant transmission values.

\section{Semantic color prior}\label{motivation}
We proposed a novel method that explores semantics for dehazing by training a CNN model to learn from training set the color distribution conditioned on a set of semantic features. This approach allows our model to infer semantic priors for recovering true scene color. The extra knowledge obtained with semantic cues is used to remedy the lack of information in a single image when the model sees similar semantics again. \\%More specifically, we exploit high level semantic features for learning semantics-related color priors for real world objects. Admittedly, some object classes can be of arbitrary radiance (\eg building or car surface can be of any color), such that color shifting resulted from different haze concentrations can seem equally plausible, making it difficult to infer the true radiance or the haze density. However, for objects of innate color priors (such as vegetation), semantic features provide cues that are much more confident and informative (\eg sky is blue and vegetations are green), as illustrated in Figure~\ref{fig0}.We find this to be particularly helpful in predicting scene radiance and illumination color.\\
\textbf{Clean image:}
\sout{\shaodi{Add in some equations and symbols to make the concept more specific. And use some of equations to describe your network if possible.} only equation I can think of is conditional probability but the model output is not probability but features and I'm scared reviewers may attack it... Added a paragraph for global features.}
Conventional methods rely on the physical haze model to restore the true color of an object. This requires an accurate estimation of the atmospheric light and transmission value (structural information), both are difficult to obtain. However, there often exists a strong correlation between a semantic class and the color distribution it exhibits (\eg vegetations are likely to have green color), as such, the true color can sometimes be predicted directly with a high confidence (see Figure~\ref{fig:intro}). The semantic features thus can offer a strong prior for the prediction of clean image which can be particularly useful when the estimation ambiguity is high (\eg very small transmission value). An example is the sky region with effectively infinite depth, in which case it is impossible to recover the true color. However, when correctly identified as sky, the image part is likely to be colors of blue hue. Although the guess is not necessarily accurate, a color distribution can nonetheless be learned and exploited to reduce the ambiguity of prediction. \sout{\todos{make sure in the experiments you show a corresponding example}}\\
\textbf{Ambient illumination:}
On the other hand, semantic context can also be useful for estimating atmospheric illumination color, the most straightforward case also being the sky regions, which often have color close to ambient light. This may in turn benefit objects with weak semantic priors (\eg cars can be of arbitrary color, but a car spatially adjacent to road or tree will likely have similar depth and transmission, and its true color may then be inferred given ambient illumination, as illustrated in Figure~\ref{fig:intro}).

In practice, however, we also observe that objects' true color and ambient illumination are mutually dependent when given hazy image as input. Consequently, we design a network to incorporate them both and allow one to refine the other. Instead of asking our network to explicitly predict illumination color, we make use of a set of global features, which may carry global contextual information related to \eg not only ambient illumination color but also global scene semantics. This non-locality allows information learned from objects of strongly confident semantic priors to propagate to other parts of the image, and benefit the true color prediction for objects with weak semantic priors.

%Here we formalize the proposed semantic dehaze approach by a probability factorization problem. Define an input haze image $I$ and its semantics $S$, for each local patch $\Omega$ we compute the clean image patch $J_\Omega$ with
%\begin{equation}
%P(J_\Omega|I) = P(S|I) \cdot P(J_\Omega|S,I)
%= P(S_\Omega|I_\Omega) \cdot P(J_\Omega|S_\Omega,I_\Omega)
%\end{equation}

%advantage of this approach, should this be in experiments part?
%We show later in experiments that the proposed model is able to leverage semantic features for robustness against severe color shifting caused by saturated ambient illumination colors or very small transmission value, and can accurately recover objects of bright color. Those are some most common irregularities that significantly impede existing methods.
%\section{Methodology}
%In this section we introduce the proposed semantic dehazing model, which consists of a semantic subnet for extracting scene semantic features, a global feature extraction subnet, as well as a pipeline which learns semantic color priors.

\begin{figure}
\centering
\includegraphics[width=\textwidth, trim={0 6cm 3cm 3cm}, clip]{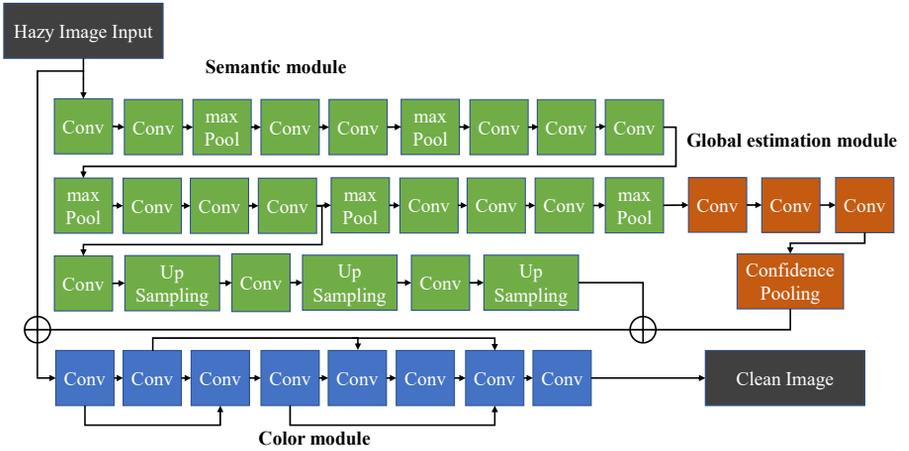}
\caption{Pipeline of the proposed model consists of a semantic module, a global estimation module and a color module (colored in green, orange, and blue respectively). Note that the outputs from both semantic module and global estimation module are up-sampled/broadcast to original images size, and concatenated with hazy image as inputs to color module.}\label{fig:architecture}
\sout{\shaodi{Redo this figure, make it more structured and easier to read.}}
\end{figure}

\subsection{Overview}
The pipeline of the proposed model is illustrated in Figure~\ref{fig:architecture}. The model takes hazy images as input and the output is the predicted clean image. Our model, following a fully convolutional design, consists of three modules: a semantic module for higher-level semantic feature extraction, a global estimation module for predicting global features, and finally a color module for inferring clean image.

\subsection{Semantic module}

%For semantic feature extraction, we exploit a well-known image classification network VGG16~\cite{simonyan2014very} that is pre-trained on and published with places365 dataset~\cite{zhou2017places}. The model, called Places365-VGG, has been extensively trained for scene classification task of 365 scene semantic categories \cite{zhou2017places}. 
For semantic feature extraction, we exploit a well-known image classification network VGG16~\cite{simonyan2014very} that is pre-trained on ImageNet dataset~\cite{ImageNet}. \sout{\todos{does the fact that it is pre-trained on image net increase the generality of the semantic classes?}} The model has been extensively trained for object recognition task over 1,000 semantic categories \cite{ImageNet}. As we only need the semantics-related features rather than the exact labeling, we remove the final dense and softmax layer of the VGG model, and use the output of its intermediate convolutional layers for semantic feature extraction. We chose VGG16 for its good performance and simplicity in design. VGG16 model has 5 multilayer blocks (which we call block 1 to block 5), each has several convolutional layers followed by one max pooling layer \cite{simonyan2014very}. Since we want to enforce the generalizability of semantic network, we do not train the VGG model to serve our dataset, \ie its weights are fixed during training. 

It is commonly observed that as information propagates through deep CNN (\eg classification and recognition models), the processed knowledge generally becomes less informative but more task-oriented. In order to balance the trade-off between information loss and the task-specificness (which in this case is scene semantics extraction), we use the final convolution layer in block 4 of VGG16 (down-sampled by a factor of 8) for local features to infer color priors. The extracted local semantic features are then up-sampled to original image size by a light-weighted three-stage subnet to accommodate the input size of color module. In each stage, the number of features is reduced by half using a convolutional layer with kernel size $3\times 3$, and the feature maps are then up-sampled by a factor of 2.

\subsection{Global estimation module}

We further exploit the semantic module for estimating a set of 32 global features \sout{for \todos{unfinished sentence}}. The key intuition behind this design is that global features may carry valuable information about ambient illumination or semantic context, which can be inferred from scene semantics. The block 5 output of the VGG16 model \sout{\todos{I am confused by block4 block5, maybe it's a CNN thing but does not appear in figs in here}} (down-sampled by a factor of 32) is sent to our global estimation module which predicts a single value ($1\times1$ size feature) for each feature map.

We adopt the confidence-weighted pooling technique proposed in~\cite{hu2017fc}. The proposed global estimation module is trained to learn from each input patch $\Omega$ a set of local features ${\cal{F}}_\Omega$, as well as the local confidence $C_\Omega$. The global features $\cal{F}$ is then obtained by averaging local features weighted by their individual confidence:

\begin{equation}
{\cal{F}} = \frac{1}{\sum_\Omega C_\Omega}\sum_{\Omega}{\cal{F}}_\Omega\cdot C_\Omega\text{.}
\end{equation}

%Hu~\etal~\cite{hu2017fc} used confidence-weighted pooling for learning the color constancy of an image. 
This pooling technique enables our model to extract global features depending on the confidence level of semantic priors on local regions, as some semantic classes may have higher significance than others (as shown in Figure~\ref{fig:intro}). The global pooling allows local features to be aggregated and broadcast to other part of image, and effectively enables image-sized receptive field in a fully convolutional architecture. \sout{\todos{any reason why?}}

To this end we build a light-weight model with four intermediate layers to learn global features. We use the first three convolutional layers (of filter size $5\times 5\times 256$, $5\times 5\times 64$ and $1\times 1\times 33$ respectively) for further feature extraction. The input to confidence pooling layer has $33$ features, where the first $32$ are the predicted local features and the last feature is the corresponding confidence. \newtodos{is this correct:"locally predicted local features", or you wanted to say just"predicted local features".}
A layer-wise softmax activation is applied on confidence for normalization purpose before pooling. The final output of global estimation module is reduced to a $1\times 1$ feature map of $32$ channels. %The experiments show a good performance of the proposed design. \sout{This design works well in our experiments.\todos{maybe: "the experiments show a good performance of the proposed design"}}

\subsection{Color module}

%inputs
The color module reads in both semantic features and global features. The global features are broadcast to original image size, at which point the hazy image and semantic and global feature maps are concatenated, as illustrated in Figure~\ref{fig:architecture}.

%difference to AODNet
The concatenated inputs are then processed by our color module. For this part, we use the architecture of AOD-Net~\cite{li2017aod} since it is the state-of-art end-to-end CNN dehazing model. However, in our case, the input contains not only hazy images, but also a $48$-channel feature map ($16$ for semantic features and $32$ for global features), and consequently, intermediate layers have more filters to process the additional input features. 
The original AOD-Net has $3$ filters for each of its $5$ convolutional layers and our modified version has $16$, $16$, $8$, $4$, $3$ filters respectively. No other modifications were done. The final output of our network is an RGB image of predicted clean scene.

\section{Experiment Evaluations}

%overview of experiments
We conduct extensive experiments, to evaluate the performance of the proposed method. We compare our model both qualitatively and quantitatively against other existing methods on both synthetic and real hazy images. We are also interested in testing under some challenging settings, e.g. non-grayscale illumination color and/or very small transmission values --- both cause estimation ambiguity. Finally, we conduct ablation study to compare our model with a baseline implementation without semantic information. 

\subsection{Datasets}\label{datasets}

%how we build our dataset
%Most previous deep learned image dehaze models were trained on indoor datasets (\eg NYU2~\cite{silberman2012indoor}) \cite{ren2016single,li2017aod,yang2018towards} for more accurate depth maps. 
Considering that the indoor settings may not be applicable for hazy scenes and indoor and outdoor images contain different semantic classes, in our experiments, we combine the labeled NYU/NYU2~\cite{silberman11indoor,silberman2012indoor} and the Cityscapes~\cite{Cordts2016cityscapes} datasets for training and testing. All images are resized to $256\times 256$ pixels to fit the input size of the pre-trained VGG16 model. We do not include Middlebury dataset for testing because Middlebury has limited depth range and different semantic classes from NYU/NYU2\newtodos{maybe also not useful semantic classes}; RGB datasets with learned depth (\eg RESIDE) are not considered for training or testing either as their depth prediction often suffers poor quality. 

\sout{\todos{"The hazy images dataset is constructed ...}} The hazy dataset used in this paper is generated from the RGB-D images following the physical model in \eqref{phyxmdl} and \eqref{trnsdef}. The Cityscapes dataset does not provide depth map, and the depth information is instead given in the form of disparity maps. As such, the depth for Cityscapes \sout{\todos{technically distance is not inv proportional to disparity, only depth is.}} is calculated inversely proportional to the disparity values, \ie the transmission is obtained from $e^{-\frac{\beta}{D}}$ where $D$ is the disparity value. To handle occlusions, we first crop the Cityscapes images \sout{\todos{do you do this only for the cityscape, if so say something like "for the cityscape dataset "?}} so the left and bottom margins with most occluded parts are removed. We then adopt the nearest neighbor assignment approach~\cite{huq2013occlusion} to fill in the missing values for remaining occluded pixels. 

The RGB-D datasets used in our experiments \newtodos{the datasets} are split for training and testing purposes as shown in Table~\ref{tab:datasetsplit}.
\sout{\todos{move the test dataset in here}}
\begin{table}[!tbp]
\caption{Breakdown of unique RGB-D images for training and testing purposes}\label{tab:datasetsplit}
\begin{center}
\begin{tabular}{|c|c|c|c|c|c|}
\hline
\multirow{2}{*}{Dataset}& \multicolumn{2}{c}{Training} & \multicolumn{2}{|c|}{Testing} & \multirow{2}{*}{Total}\\
\cline{2-5}
& train & validation & \textbf{testsetA} & \textbf{testsetB} & \\\hline
NYU & 1524 & 152 & \multicolumn{2}{|c|}{608} & 2284 \\\hline
NYU2 & 969 & 96 & \multicolumn{2}{|c|}{384}  & 1449\\\hline
Cityscapes & 2575 & 400 & 1525 & 0 & 4500\\\hline
Total & 5068 & 648 & 2517 & 992 & 8233\\
\hline
\end{tabular}
\end{center}
\end{table}
The training and testing hazy images are synthesized using the RGB-D images, random illumination colors and different haze coefficients.
As real world haze may exhibit non-gray colors, we sample the illumination color from HSV color space. %Due to the fact that real world illumination colors are often bright and unlikely to be extremely saturated, 
In our experiments we use a lightness\sout{\todos{lightness term is not defined}} value (\ie the value of V in HSV) from $U(0.6,1)$ and saturation from $U(0,0.5)$ (where $U$ stands for uniform distribution). The hue is sampled in full range of $U(0,1)$ since there is no prior knowledge on it. For NYU and NYU2 dataset, the haze coefficient $\beta$ is uniformly in $\{0.1, 0.2, 0.3, 0.4\}$ (unit is $m^{-1}$) and for Cityscapes it is uniformly in $\{5, 7.5, 12.5, 20\}$ (unit is the disparity unit) \sout{\todos{This paragraph is still not clear! Please make reference to eq. 1 when you provide the values. Why $\beta \in {0.1 \dots 0.4}$ for NYU and $\beta \in {5 \dots 20}$ for cityscape. Make sure you mention the values are sampled from a uniform distribution. The experiments need to be reproducible, so the description of the datasets need to be clear.}\todos{I mentioned later that for each RGB-D image and each $\beta$ we generate one image per epoch, do I still say it's uniformly sampled}}.

From each of the 2,517 RGB-D images in the test set we generate $4$ hazy images using different $\beta$ and a random illumination color as described above. We call this \textbf{testsetA}. Since NYU/NYU2 datasets are widely used for quantitative evaluation in the single image dehaze literature~\cite{li2017reside,ancuti2016dhazy}, % and most of the existing CNN methods were trained and tested on~\cite{ren2016single,li2017aod,yang2018towards}. 
for better comparability, we construct another \textbf{testsetB}, which is synthesized using only test images from NYU and NYU2 in Table~\ref{tab:datasetsplit} and contains a totality of 3,968 hazy images. For each RGB-D image and $\beta$, we generate one hazy image with grayscale illumination color (illumination color is from HSV space with $S=0$ and $V$ from $U(0.6,1)$). 

\newtodos{what about the real hazy images dataset}

\subsection{Training setup} 
\sout{\todos{here you need a sentence summarizing the training procedure. e.g. ``We trained our network on...''  }}
%training routine and hyper parameters
We train our model on the RGB-D training set using haze generation scheme described above. \newtodos{ above you mainly specified the test dataset. Here you need to say something. We construct our training set in a similar way as testsetA.}
The training inputs are shuffled for each epoch and batched to size of $8$. The model is trained with Adam optimizer~\cite{Kingma2014AdamAM} at learning rate of $1e-4$ and default momentums. After each epoch we evaluate the model on the validation set. An early stop criterion applies when training loss has converged and the best validation loss has not improved over $7$ consecutive epochs.

%train 'on the fly'
Given the massive amount of used RGB-D images, instead of keeping a static hazy dataset like we did before with \textbf{testsetA} and \textbf{testsetB}, we generate the training and validation hazy images on the fly. \newtodos{is this only for the training set? check the change} During every epoch we loop over all RGB-D images and for each image with each pre-defined haze coefficient, generate one pair of hazy/clear scene with randomly sampled illumination colors. In this way, we increase the diversity of training samples by not reusing pre-stored hazy images.%, when also reducing the overhead to generate millions of hazy images before training.

\subsection{Quantitative evaluation}

We compare our model with the state-of-art methods described in section \ref{sota} on synthetic hazy images. For that we use the mean squared error (MSE), peak signal to noise ratio (PSNR) and structural similarity (SSIM) metrics. Since \cite{yang2018towards} did not publicize their source code or trained model, we do not include it for comparison. 
%For fair comparison, all methods are tested on the same sets of images mentioned before \newtodos{I think this sentence should be removed or expanded}.

The results listed in Table~\ref{tab:color} and \ref{tab:nyugray} are evaluated on \textbf{testsetA} and \textbf{testsetB}, respectively. Given the large number of test images, we use the default optimal parameters reported in the corresponding papers or implementations for comparing state-of-art methods. In both Table~\ref{tab:color} and \ref{tab:nyugray}, our method produces significantly better results than existing methods, suggesting that the semantic prior is a powerful tool for image dehazing. 

Table~\ref{tab:color} shows that our method is robust against the estimation ambiguity introduced by different illumination settings. Since the illumination is not on grayscale, handcrafted priors such as CAP are violated. Previous CNN methods also do not perform well under this settings, because (a) they are light-weighted model with limited learning capacity as they are designed to learn only low-level features and (b) they are trained on less challenging dataset with only grayscale haze scenes\newtodos{are you sure about the less hazy part}. On \textbf{testsetB}, however, CNN approaches generally have better performance than hand-crafted priors (except for DCP~\cite{he2011single}). Interestingly, while Haze-lines~\cite{berman2016non,7951489} has the lowest score on PSNR metric, it has the second best score on SSIM in Table~\ref{tab:color}. This is in line with our later observation where Haze-lines shows noticeably better image quality than other existing methods \newtodos{I never heard of "previous-arts" in the context of the state-of-the-art methods. In general papers avoid short forms, \ie it's}. %Note that our testset contains more than 8k hazy images, and evaluation on the entire testset would take an exceedingly long time to finish. As such we test some existing methods (marked with *) in Table~\ref{tab:color} on only a random subset of the test images. To ensure the accuracy of evaluation, we increase the number of images $n$ until the average testing loss $\mu$ converges to a reasonable point where standard deviation $\sigma<0.001\sqrt{n}\mu$. 

\newtodos{any comments on the values they report in the paper? (AOD)}
\begin{table}[tbp]
\caption{Comparison with the state-of-art methods over SSIM, PSNR (larger is better) and MSE (lower is better) metrics on synthetic hazy images in our \textbf{testsetA} with color scale haze. CNN models are listed on the left and handcrafted priors methods are listed on the right.}\label{tab:color}
\begin{center}
\begin{tabular}{|c|*4{|c}|*4{|c}|}
\hline
Metrics & Ours & DehazeNet\cite{cai2016dehazenet} & MSCNN\cite{ren2016single} & AOD-Net\cite{li2017aod} & DCP\cite{he2011single} & CAP\cite{zhu2015fast} & Haze-lines\cite{berman2016non,7951489}\\
\hline\hline
SSIM & \bf{0.9018}  & 0.5829  & 0.6487  & 0.5159  & 0.6329 & 0.5675 & 0.6598\\
MSE & \bf{0.0020} & 0.0311 & 0.0255 & 0.0337 & 0.0243 & 0.0320 & 0.0405\\
PSNR & \bf{28.195} & 16.735 & 17.075 & 15.537 & 17.063 & 15.752 & 15.515 \\
\hline
\end{tabular}
\end{center}
\end{table}
\sout{Table~\ref{tab:nyugray} shows the results. \todos{comment those results. e.g. `` improvement over the table 2'', seems that there is enough space to join both tables on the same lines (next to each other) This will facilitate the comparison.}}
\begin{table}[!btp]
\caption{Comparison with the state-of-art methods over SSIM, PSNR (larger is better) and MSE (lower is better) metrics on synthetic hazy images in \textbf{testsetB} with grayscale haze. CNN models are listed on the left and handcrafted priors methods are listed on the right.}\label{tab:nyugray}
\begin{center}
\begin{tabular}{|c|*4{|c}|*4{|c}|}
\hline
Metrics & Ours & DehazeNet\cite{cai2016dehazenet} & MSCNN\cite{ren2016single} & AOD-Net\cite{li2017aod} & DCP\cite{he2011single} & CAP\cite{zhu2015fast} & Haze-lines\cite{berman2016non,7951489}\\
\hline\hline
SSIM & \bf{0.9024} & 0.8140 & 0.7621 & 0.8027 & 0.8212 & 0.7744 &  0.7146 \\
MSE & \bf{0.0027} & 0.0158 & 0.0217 & 0.0208 & 0.0157 & 0.0215 & 0.0290 \\
PSNR & \bf{27.083} & 20.291 & 18.229 & 18.103 & 19.391 & 17.920 & 16.722\\
\hline
\end{tabular}
\end{center}
\end{table}

\subsection{Qualitative evaluation}

In this section we qualitatively compare the results of the proposed model and state-of-art methods over a collection of synthetic and real world hazy images. The images used in this evaluation are taken from our \textbf{testsetA} \sout{\todos{which one? (NYU+CityScape)+color haze?}} as well as real world scenes from the Internet and Foggy Driving dataset \cite{foggy}. Given the semantic-aware nature of our model, the real world scenes for comparison are selected to contain similar semantic classes to the training set. \newtodos{are you planing to release the real world set?}

\begin{figure}
\hspace{-1.5cm}
\includegraphics[page=2, width=1.25\textwidth, trim={0cm 8cm 0cm 8cm}, clip]{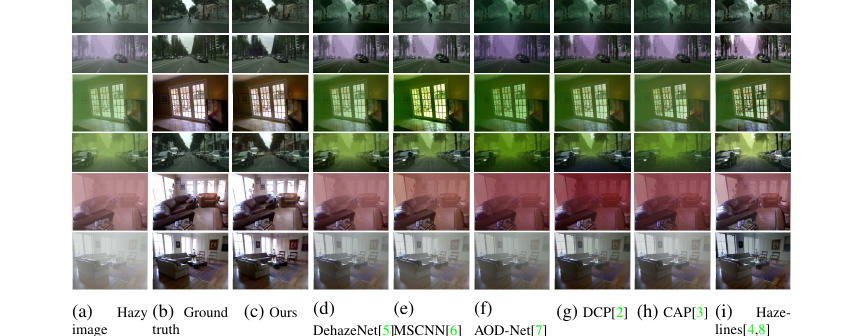}
    \caption{Visual comparison with the state-of-art methods on our \textbf{testsetA}. Images are resized for viewing purpose.}\label{fig:visual1}
\end{figure}

\begin{figure}
\hspace{-1.5cm}
\includegraphics[page=3, width=1.25\textwidth, trim={0cm 2.5cm 0cm 2.5cm}, clip]{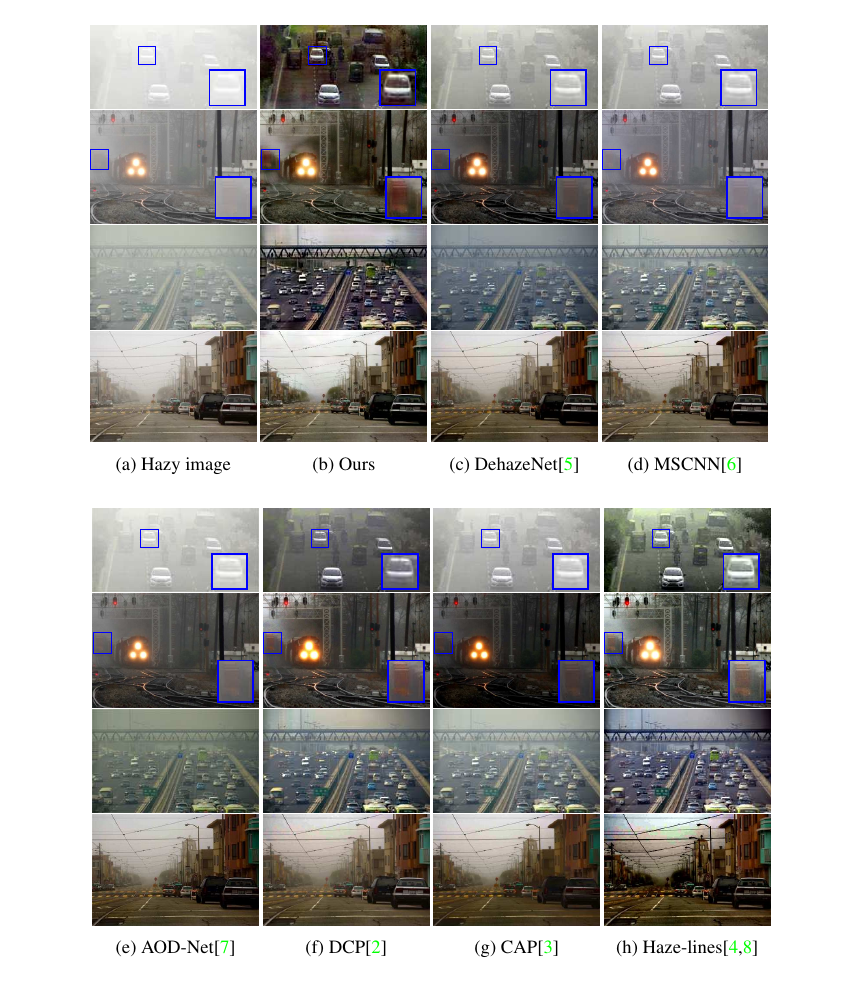}
    \caption{Visual comparison with the state-of-art methods on real world hazy images of urban scenes from Internet and \cite{foggy}.}\label{fig:visual2}
\end{figure}

\begin{figure}
\hspace{-1.5cm}
\includegraphics[page=4, width=1.25\textwidth, trim={0cm 8.5cm 0cm 8.5cm}, clip]{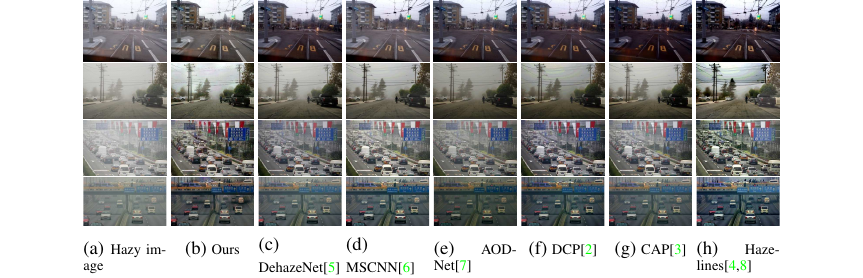}
\caption{Visual comparison with the state-of-art methods on real world hazy images of
urban scenes from Internet and \cite{foggy}.}\label{fig:hazy4}
\end{figure}

Figure~\ref{fig:visual1}-\ref{fig:hazy4} show that our model can recover the scene under very strong haze and restore plausible color to objects indistinguishable to human eyes. One notable example is the sky region in Cityscapes dataset shown in Figure~\ref{fig:visual1}. Although the sky has very small transmission values, \ie it suffers from severe color shift, our model is still able to recover it naturally. In Figure~\ref{fig:visual2}, the images with medium and strong haze levels are extrapolated properly, and the brightness and color balance restored. The trees in the top row appear green and our method produces less artifacts than DCP and Haze-Lines. 

On the other hand, existing methods are susceptible to over-saturation, especially when illumination color is not on grayscale. A noticeable example is the predicted image by DCP~\cite{he2011single} in Figure~\ref{fig:visual1}, where the saturation is apparently increased, causing color shifts even more from the clean scene seen in ground truth images. Among existing methods, Haze-lines~\cite{berman2016non} has arguably the best visual quality in Figure~\ref{fig:visual2}, when other existing methods fail to properly restored white balance or fully lift the haze in the image. 

This gap in visual quality is mainly because the semantic approach enables our model to understand what is being imaged, which provides informative priors that are otherwise impossible to learn from lower-level features alone.

\subsection{Ablation study}

To further validate the proposed semantic approach, we compare our model with a baseline that receives no semantic features. In this ablation study we aim to exam the semantic dependency of our method and how that would impact performance. To this end, we remove the semantic-related features by isolating the color module from the rest of the pipeline, \ie both semantic module and global estimation module are dropped out from the design. 

Since the semantic and global estimation modules may learn to predict illumination-related features as well as semantic features, here we use the same dataset introduced before, but with ground truth RGB ambient illumination color as additional input when training and testing. We concatenate the hazy image with illumination color to form a 6-channel input (as opposed to the 51-channel input of original color module). To ensure that this does not undermine the model's learning capacity, the number of filters in the first convolutional layer is increased to $24$ so the baseline has slightly more parameters than the original color module. Otherwise there is no difference between the baseline and our original color module. For fair comparison we train the baseline model with the same dataset and training routine described before.

\begin{table}[!t]
\caption{Comparison of baseline and the proposed method over different metrics on \textbf{testsetA}. Both models are trained and tested on the same dataset.\label{tab:ablation}}
\begin{center}
\begin{tabular}{|l|c|c|c|c|}
\hline
Method & SSIM & MSE & PSNR \\
\hline\hline
Ours & \bf{0.9024} & \bf{0.0020} & \bf{27.0827} \\
Baseline & 0.7762 & 0.0061 & 23.2918 \\
\hline
\end{tabular}
\end{center}
\end{table}

Table~\ref{tab:ablation} shows the performance of baseline over \textbf{testsetA}. \newtodos{are you using the testsetA in here?} The performance varies considerably from the baseline to our semantic approach even when we provide additional ground truth illumination color to the former. Note that although the baseline shares the same layer architecture with AOD-Net~\cite{li2017aod}, it still wins by a large margin over our test set. This is because (a) the baseline is trained on more challenging dataset with non-grayscale ambient illumination (b) the baseline has more filters (parameters) than AOD-Net and (c) ground truth illumination is known to the baseline model but not to AOD-Net.

\section{Conclusion and discussions}

In this paper we introduced a semantic approach towards single image dehazing. We are the first to explicitly exploit semantic features for learning semantic priors which are used to provide informative priors for estimating underlying clean scene. We achieved state-of-the-art performance on synthetic hazy images and our model is able to accurately recover clean scene under strong estimation ambiguity, \eg strong haze and semi-saturated ambient illumination, with learned semantic priors. 

Due to the difficulty for acquiring real world training data, our dataset contains only indoor scenes and outdoor road scenes, which implies a deficit for semantics of general real world objects as well as their corresponding real colors. Therefore, it is challenging to learn semantic-color priors for real world outdoor objects that are not seen during training, which means we cannot show our model can generalize well to natural outdoor scenes until relevant datasets are made available. This limitation is on the published data rather than on our methodology, and the generalizability is traded for much higher accuracy seen in experiments. In future we will improve this by training our model with a wider range of images.

%\clearpage

\section{Acknowledgement}
We would like to thank Christos Sakaridis for his careful review of our paper which helped improve its quality.

\end{document}